\documentclass[twoside]{article}
\usepackage[T1]{fontenc}
\usepackage[accepted]{aistats2024}

\usepackage[centerfigures,citations,authoryear,commands,enumerate,environments,theorems,lmrscinnershape]{AVT}
\usepackage{array}
\newcolumntype{P}[1]{>{\centering\arraybackslash}p{#1}}
\usepackage{booktabs}
\usepackage{stfloats}
\usepackage[flushleft]{threeparttable}
\usepackage{subcaption}
\usepackage{multirow}
\usepackage{algorithm}
\usepackage{algorithmicx}
\usepackage{algpseudocode}
\usepackage{xfrac}
\usepackage{xcolor}
\definecolor{lightblue}{RGB}{66, 170, 245}
\newcommand{\LeftComment}[1]{\Statex \textcolor{lightblue}{\textbackslash\textbackslash \space \space \space #1}}
\usepackage{tikz}

\bibliography{GPMP}

\runningtitle{A Unifying Variational Framework for Gaussian Process Motion Planning}
\runningauthor{L. Cosier, R. Iordan, S. Zwane, G. Franzese, J. T. Wilson, M. P. Deisenroth, A. Terenin, Y.~Bekiroğlu}
\begin{document}
\twocolumn[
\aistatstitle{A Unifying Variational Framework for \\ Gaussian Process Motion Planning}

\aistatsauthor{Lucas Cosier\textsuperscript{\ensuremath{*1,2}}
\And Rares Iordan\textsuperscript{\ensuremath{*1}} \And Sicelukwanda Zwane\textsuperscript{\ensuremath{1}} \And Giovanni Franzese\textsuperscript{\ensuremath{6}}}

\aistatsauthor{James T. Wilson\textsuperscript{\ensuremath{3}} \And Marc Peter Deisenroth\textsuperscript{\ensuremath{1}} \And Alexander Terenin\textsuperscript{\ensuremath{3,4,5}} \And Yasemin Bekiroglu\textsuperscript{\ensuremath{1,7}}
}

\aistatsaddress{\textsuperscript{\ensuremath{1}}University College London\qquad\textsuperscript{\ensuremath{2}}ETH Zürich\qquad\textsuperscript{\ensuremath{3}}Imperial College London\qquad\textsuperscript{\ensuremath{4}}University of Cambridge\\\textsuperscript{\ensuremath{5}}Cornell University\qquad\textsuperscript{\ensuremath{6}}TU Delft \qquad\textsuperscript{\ensuremath{7}}Chalmers University of Technology}
]

\begin{abstract}
To control how a robot moves, motion planning algorithms must compute paths in high-dimensional state spaces while accounting for physical constraints related to motors and joints, generating smooth and stable motions, avoiding obstacles, and preventing collisions. A motion planning algorithm must therefore balance competing demands, and should ideally incorporate uncertainty to handle noise, model errors, and facilitate deployment in complex environments. To address these issues, we introduce a framework for robot motion planning based on variational Gaussian processes, which unifies and generalizes various probabilistic-inference-based motion planning algorithms, and connects them with optimization-based planners. Our framework provides a principled and flexible way to incorporate equality-based, inequality-based, and soft motion-planning constraints during end-to-end training, is straightforward to implement, and provides both interval-based and Monte-Carlo-based uncertainty estimates. We conduct experiments using different environments and robots, comparing against baseline approaches based on the feasibility of the planned paths, and obstacle avoidance quality. Results show that our proposed approach yields a good balance between success rates and path quality.
\end{abstract}

\section{Introduction}

\begin{table}[b!]
\raggedright\small\urlstyle{same}\textsuperscript{\ensuremath{*}}Equal contribution.\\Code repository: \textsc{\url{https://github.com/luke-ck/vgpmp}}.\\Video available in supplementary material: see repository.
\vspace*{0.8125ex}
\end{table}

Motion planning refers to the process by which a robot finds a path of motion from a start to a goal state. 
This path should be collision-free, satisfy constraints such as joint or torque limits, maintain a desired end-effector orientation, and fulfill other task-specific requirements.
The path should also be smooth, avoiding abrupt or otherwise sharp motions.
To compute the path, an algorithm needs to be able to handle potentially high-dimensional robot configuration spaces. 
This may involve uncertainty, due to inherent sensor noise, external disturbances, and other effects. 
The algorithm also needs to cope with complex environment~dynamics.

Modern motion planning algorithms handle these challenges using different mathematical principles.
\emph{Sampling-based} planners, such as rapidly-exploring random trees \cite{kuffner-lavalle, cheng2002resolution, rrtstar, berenson2011task}, work by exploring the space of possible trajectories via stochastic processes, such as space-filling trees.
\emph{Optimization-based} planners \cite{Ratliff, schulman2013, schulman2014, Zucker, stomp, gpmp, JointMotionGrasp} calculate a motion plan by solving a constrained optimization problem defined on the space of possible paths. \emph{Probabilistic-inference-based} planners \cite{gpmp2, attias, Toussaint-storkey, Toussaint-goerick, gvi, iGPMP2} reinterpret motion planning as a Bayesian inference problem over spaces of paths, enabling motion plans to be computed using inference algorithms. 

Once a path is computed, one must ensure it can be executed successfully on the physical robot.
Using probabilistic representations to handle uncertainty is becoming increasingly prominent in a number of related areas, such as imitation learning. 
For instance, \emph{probabilistic motion primitives} \cite{promps} parameterize Gaussian trajectory distributions to learn from expert demonstrations, and provide related functionality such as blending trajectories together.
\textcite{ewerton} extend this by adding a mixture model to handle distinct candidate trajectories, with uncertainty around each trajectory.
\textcite{sgpmp} integrate expert demonstrations to create stochastic motion planning primitives using energy-based-models.
These developments highlight the usefulness of probabilistic representations to address uncertainty in related robotics problems.

In this work, we study probabilistic-inference-based motion planning using Gaussian processes.
We (i) introduce a unifying \emph{variational Gaussian process motion planning (vGPMP)} framework, which represents the motion planning problem as a \emph{variational inference problem}---an optimization problem on a suitable space of probability distributions.
This problem is parameterized by a set of \emph{waypoints}, which describe locations through which the motion plans should go through, on average: these waypoints can be optimized to find motion plans for a given task.
Furthermore, we (ii) show, using this variational-inference-based viewpoint, that Gaussian process motion planners are stochastic extensions of optimization-based planners.
As a result, Gaussian process motion planners implemented in this manner share the same capabilities as optimization-based planners, and additionally provide uncertainty estimates due to their stochastic construction, which can be propagated downstream.

Our framework makes it possible to directly incorporate both hard and soft constraints, including joint limits and collision constraints, into the motion planning objective as well as task-independent properties, such as smoothness.
We enable both interval-based and Monte-Carlo-sampling-based quantification of uncertainty, each in a scalable manner. 
The resulting framework augments and simplifies preceding Gaussian-process-based motion planners \cite{gpmp2, gpmp2-graph}, giving practitioners a straightforward-to-implement way to construct motion plans with uncertainty.

\section{Motion Planning using Gaussian Processes} \label{sec:gps}

Gaussian processes (GPs) are random functions $f: \c{X} \-> \R$, where the output value of the function at any finite set of input variables $\v{x} \in \c{X}^n$ follows a Gaussian distribution.
A Gaussian process is characterized by a mean function $\mu: \c{X} \-> \R$ and a positive semi-definite kernel $k: \c{X} \x \c{X} \-> \R$, such that, if $f \~ \c{GP}(\mu, k)$, then $\v{f} = f(\v{x}) \~ \c{N}(\v\mu, \m{K}_{\v{x}\v{x}})$ with mean $\v\mu = \mu(\v{x})$ and covariance $\m{K}_{\v{x}\v{x}} = k(\v{x}, \v{x})$.
Let $(\v{x},\v{y})$ be the data, and to ease notation assume the prior mean is $\mu = 0$.

If we consider a Gaussian likelihood $p(\v{y}\given \v{x}) = \c{N}(f(\v{x}), \sigma^2\m{I})$ where $\sigma^2 > 0$ is the noise variance, then the conditional distribution of the random function $f$ given the training dataset $(\v{x},\v{y})$ is also a Gaussian process.
Letting $\m{K}_{(\.)\v{x}} = k(\.,\v{x})$, we have $f\given\v{y} \~\c{GP}(\mu_{f\given\v{y}},k_{f\given\v{y}})$ where
\[
\mu_{f\given\v{y}}&= \m{K}_{(\.)\v{x}}(\m{K}_{\v{x}\v{x}} + \sigma^2 \m{I}_n)^{-1}\v{y}
&\\
k_{f\given\v{y}} &= k(\.,\.') - \m{K}_{(\.)\v{x}}(\m{K}_{\v{x}\v{x}}+ \sigma^2\m{I})^{-1}\m{K}_{\v{x}(\.')}
.
\]
To train the GP, the hyperparameters of the kernel need to be learned.
This is typically done by maximizing the marginal likelihood using standard gradient-based optimization methods \cite{rasmussen-williams}.

In the context of motion planning, the input space will always represent time, namely $\c{X} = \c{T} = \R$, and the output space will represent unconstrained joint values and possibly their derivatives, which we write as $\Theta = \R^d$.
The GPs we consider will therefore be random functions $f: \c{T} \to \Theta$, and the data will represent joint values and derivatives at a particular time.

\subsection{Gaussian Process Motion Planning}
\label{sec:gpmp2}

By modeling the distribution of feasible trajectories as functions that map time to robot states, GPs provide a principled approach for formulating the motion planning problem through the lens of probabilistic inference.
Gaussian processes were introduced to motion planning by \textcite{gpmp2}, who propose to choose a GP prior which is given by a \emph{linear time-varying stochastic differential equation}, namely
\[
\label{eqn:lti-sde}
\d\v{f}(t) &= (\m{A}(t)\v{f}(t)  + \v{u}(t))\d t + \m{F}(t) \d\m{w}(t)
\]
where $\m{A}(t)$ and $\m{F}(t)$ are time-varying matrices, $\v{u}(t)$ is the control input to the system, and $\m{w}(t)$ is a Wiener process. 
The solution of this equation defines the GP prior.
The primary motivations for this choice are that it (i) encodes spatial smoothness, and (ii) gives rise to structured covariance matrices, which are block-tridiagonal because the stochastic differential equation (SDE) possesses a Markov property. 
\textcite{gpmp2} show that this accelerates computation and enables fast maximum a posteriori inference.

To complete the model, \textcite{gpmp2} introduce a likelihood which encourages the robot to avoid collisions and constraint violations. 
Let $\f{k}_{\f{fwd}} : \Theta \-> \Psi$ be the forward kinematics function, where $\Psi$ represents joint locations and orientations. 
Given the forward kinematics output, following \textcite{Ratliff}, they approximate the robot's body by a set of $s$ spheres, and calculate a signed distance field for each sphere: let $\f{sdf}_s : \Psi \-> \R^s$ be the function representing this calculation.
Let $\f{h}_\eps(x) = \max(-x + \eps, 0)$ be the hinge loss, where $\eps > 0$ is called the \emph{safety distance} parameter.
Let $\f{c}$ be a linear penalty function for values exceeding a given joint limit threshold.
By convention, $\f{h}_\eps$ and $\f{c}$ act component-wise on vectors.
Let $\norm{\v{x}}_{\m\Sigma}^2 = \v{x}^T\m\Sigma^{-1}\v{x}$ be the Mahalonobis norm, and let $\m\Sigma_{\f{obs}}$ and $\m\Sigma_{\f{c}}$ be diagonal matrices.
With this notation, \textcite{gpmp2} propose the likelihood $p(\v{e}\given\v{f})$ proportional to
\[
\!
\exp\del[3]{-\frac{1}{2} \del[2]{\,\ubr{\norm{\f{h}_\eps(\f{sdf}_s(\f{k}_{\f{fwd}}(\v{f})))}_{\m\Sigma_{\f{obs}}}^2}_{\t{collision term}} + \ubr{\norm{\f{c}(\v{f})}^2_{\m\Sigma_{\f{c}}}}_{\t{soft constraint term}}\,}\!}
\label{eq:likelihood form}
\!
\]
where $\v{e}$ represents the probability of \emph{collision-free events with motion constraints}.
From here, \textcite{gpmp2} approximate the posterior distribution as follows: (i) apply maximum a posteriori inference, deriving efficient algorithms which use the SDE's Markov property to recast the resulting optimization objective as inference in a sparse factor graph.
Once such a trajectory is obtained, (ii) apply Laplace approximations to the posterior distribution, and (iii) interpolate joint configuration in-between the states at times $t_1,..,t_T \in \c{T}$.
At run-time, the output values of $\v{f}$ are clamped to enforce constraints.
In total, the computations performed produce an optimal trajectory $\v{f}^*$, and a Gaussian distribution representing trajectories around it that quantifies uncertainty.

\section{Variational Gaussian Process Motion Planning}

Computing motion plans using GPs involves balancing multiple competing factors, including obstacle avoidance, smoothness, and conformity to joint limits.
These factors depend simultaneously on the motion planner's hyperparameters, which must be tuned, and on the formulation of the objective itself.
To provide practitioners with precise control over what the planner produces and enable them to modify it as needed to introduce other objectives for specific tasks, we develop a unifying mathematical and algorithmic formalism for Gaussian process motion planning.

\subsection{Sparse Gaussian Processes as a Unifying Framework}

From a probabilistic inference perspective, one of the challenges in Gaussian Process motion planning is the need to accommodate non-Gaussian likelihoods.
This means that the motion planning posterior, unlike the prior, is not a GP, which means one cannot perform exact probabilistic inference.
One must therefore use numerical methods to extract means, covariances, and Monte Carlo samples from the posterior distribution.

\emph{Sparse Gaussian Processes} \cite{hensman2013gaussian, hensman2015, titsias} are a state-of-the-art method for applying GPs in settings with non-Gaussian likelihoods.
They approximate the exact non-Gaussian posterior with the best Gaussian approximation, in the sense of Kullback--Leibler divergence.
Originally proposed to tackle scalability issues which arise when working with large datasets, they provide a broad and flexible framework for designing approximate probabilistic inference schemes using GPs.
We now apply these notions to design a unifying framework for  motion planning.

We start with an \emph{arbitrary} Gaussian process prior $f \~ \c{GP}(\mu,k)$, generalizing the framework of \textcite{gpmp2} beyond stochastic differential equations.
Thus, for any collection of \emph{times} $\m{T}= [t_0, t_1,\dots,t_n]$, the random vector evaluated on this subset $\v{f}= [\v{f}(t_0), \v{f}(t_1), \dots, \v{f}(t_n)]$ follows a joint Gaussian distribution whose mean and covariance are determined by $\mu$ and $k$. 
Simplifying, suppose that we are also given a likelihood $p(\v{e}\given\v{f})$, which incorporates constraints and collision avoidance, such as the one presented in \Cref{sec:gpmp2}---we will discuss likelihoods later in \Cref{sec:likelihoods}.
These ingredients specify the exact posterior distribution, which is non-Gaussian.

\begin{figure*}
\begin{subfigure}[b]{0.49\textwidth}
\includegraphics[height=4cm, trim=15pt 20pt 0pt -5pt]{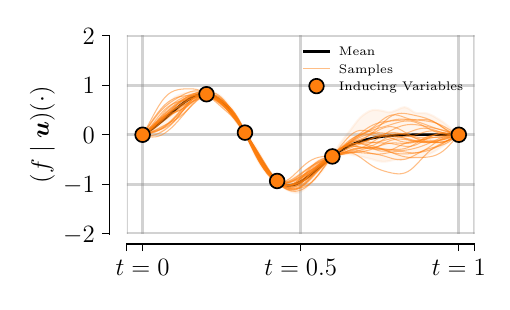}
\vspace*{1ex}
\caption{Sparse Gaussian process}
\label{fig:sparse gp}
\end{subfigure}
\hfill
\begin{subfigure}[b]{0.49\textwidth}
\includegraphics[height=3.5cm]{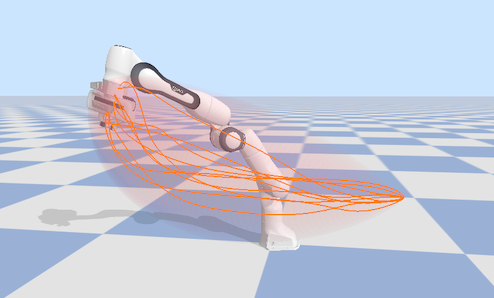}
\vspace*{1ex}
\caption{Motion plans via GP samples}
\label{fig:motion plans}
\end{subfigure}
\caption{Intuitive illustration of representing motion paths with GPs. (\subref{fig:sparse gp}) Sparse GP posterior for 1 joint with inducing variables $\v{u} = [ \v{u}_{\f{c}}, \v{u}']$ denoted by ‘$\circ$’, where $\v{u}_{\f{c}} = \left[\v{\theta}_0, \v{\theta}_1\right]$ and $\v{u}'$ represent the $M=4$ inducing locations $\v{z} \in \c{T}$ subject to the learning process, shown in-between $t=0$ and $t=1$. We interpret the inducing points as \emph{waypoints} through which the random motion plans travel. (\subref{fig:motion plans}) Candidate end-effector trajectories for the Franka Panda robot and associated uncertainty, which is illustrated as the orange shaded region.}
\label{fig:gp-posterior-samples}
\end{figure*}

To handle non-Gaussianity, following the sparse GP framework \cite{snelson-ghahramani, quinonero-candela, titsias}, we introduce the \emph{variational approximation} $f\given\v{u}$, which is a GP conditioned on the event $f(\v{z}) = \v{u}$, where $\v{z} \in \c{T}^m$ are a set of locations called the \emph{inducing points}, and $\v{u}$ are called the \emph{inducing variables}. Let $p(\v{u}) = \c{N}(\v{0}, \m{K}_{\v{z}\v{z}})$ be the prior at the $\v{z}$-locations. Further, let $q(\v{u}) \~\c{N}(\v\mu_{\v{u}},\m\Sigma_{\v{u}})$ be the distribution of the $\v{u}$-values, which we assume to be a free-form multivariate Gaussian whose mean and covariance $\v\mu_{\v{u}}$ and $\m\Sigma_{\v{u}}$ will be learned using optimization.
The gives $f\given\v{u} \~ \c{GP}(\mu_{f\given\v{u}}, k_{f\given\v{u}})$ with 
\[
\mu_{f\given\v{u}} &= \mu(\.)  + \m{K}_{(\.)\v{z}}\m{K}_{\v{z}\v{z}}^{-1} (\v{u} - \v\mu_{\v{u}})
&\\
k_{f\given\v{u}} &= k(\.,\.^{'}) - \m{K}_{(\.)\v{z}}\m{K}_{\v{z}\v{z}}^{-1}\m{K}_{\v{z}(\.)} 
\]
To learn the parameters of $q(\v{u})$, we minimize the Kullback--Leibler divergence between $f\given\v{u}$ and the true posterior.
One can show \cite{Matthews} that this is equivalent to maximizing the \emph{evidence lower bound} (ELBO)
\[
\label{elboexp}
\E_{q(\v{u})} \E_{p(\v{f}\given\v{u})} \log p(\v{e}\given \v{f}) - D_{\f{KL}}(q(\v{u}) \from p(\v{u}))
\]
which defines our general framework for probabilistic-inference-based motion planning. 
This objective includes an expectation: since sampling from the involved distributions is tractable, it can be optimized using stochastic optimization algorithms.
We use standard gradient-based techniques because they are simple, widely-used throughout machine learning, and work in practice: algorithmic details are given in \Cref{app:vgpmp_algorithm}.

The inducing points $\v{u}$ can be understood as \emph{waypoints} through which the robot should move: this is illustrated graphically in \Cref{fig:gp-posterior-samples}.
By optimizing the ELBO, the motion planning algorithm must therefore find waypoints which ensure the stochastic trajectories comply with the necessary requirements.
To understand how to assure this, we study (i) how to specify the likelihood $p(\v{e}\given \v{f})$ to incorporate these and other factors.
We then study (ii) how to extract uncertainty for use by upstream systems. 

\subsection{Incorporating Constraints, Smoothness, and Collision Avoidance}
\label{sec:likelihoods}

A motion planner must generate plans adhering to various constraints, such desired start-and-end-states and collision-avoidance. 
In the variational Gaussian Process motion planning framework, this involves integrating information into the posterior distribution using conditional probability, typically through the likelihood $p(\v{e}\given\v{f})$.
We now describe how this works for a number of different forms of information.

\paragraph{Joint Limits and Inequality-based Constraints}

To respect the robot's joint limits, we need to incorporate an inequality constraint into the GP model.
Previous, non-variational Gaussian Process motion planning approaches \cite{gpmp2} generally enforce such constraints softly by adding an extra term to the optimization objective.
This does not guarantee the constraints are satisfied, which necessitates post-processing.
The most natural alternative, which guarantees constraints to be satisfied, would be to modify the prior using conditional probability, but this unfortunately breaks Gaussianity of the GP.
We therefore instead propose to leverage the variational formulation and enforce constraints using a bijective non-linear transformation.
For box constraints of the form 
\[
\theta_i^{(\min)} \leq \theta^{\vphantom{()}}_i \leq \theta_i^{(\max)}
\]
this can be done by passing the GP through a scaled and shifted sigmoid function $\sigma : \Theta \-> \Theta_{\f{c}}$, where $\Theta_{\f{c}}$ is constrained joint space. 
This amounts to simply replacing $\v{f}$ with $\sigma(\v{f})$ in the likelihood, and returning motion plans based on the transformed GP sample paths, which are guaranteed to satisfy all joint constraints.
One can similarly handle other constraints which map bijectively into a box, such as triangular constraints.
This straightforward solution highlights the value of a variational formulation, which makes such aspects computationally simple to~add.

\paragraph{Start States, Goal States, and Equality-based Constraints}

To ensure our trajectories pass through the (unconstrained) start and goal states $\v\theta_0, \v\theta_1$ at times $t=0$ and $t=1$, we condition the variational process $f\given\v{u}$ on the events $f(0) = \v\theta_0$ and $f(1) = \v\theta_1$.
This is done by partitioning the inducing variables as $\v{u} = [ \v{u}_{\f{c}}, \v{u}']$, where $\v{u}_{\f{c}}$ are the inducing variables representing start and goal states, and $\v{u}'$ are the remaining inducing variables.
We then simply set $\v{u}_{{\f{c}}}$ to the values needed, and exclude them from optimization.
We illustrate this in \Cref{fig:gp-posterior-samples} where $\v{u}_{\f{c}}$ are set to $[0, 0]$ at $t=0, t=1$ in the 1D case. 
This approach enables one to handle general equality-based~constraints.

\paragraph{Collision Avoidance, Velocity Limits, and Additional Soft Constraints}

\begin{figure*}[t]
\begin{tikzpicture}
\node at (0,0) {\includegraphics[height=3.625cm]{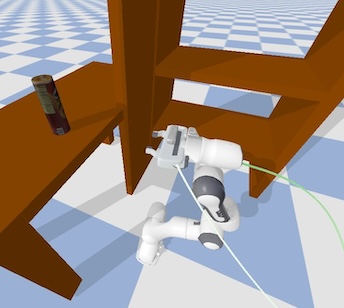}};
\node at (4.25,0) {\includegraphics[height=3.625cm]{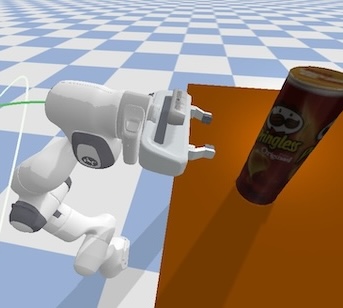}};
\node at (8.5,0) {\includegraphics[height=3.625cm]{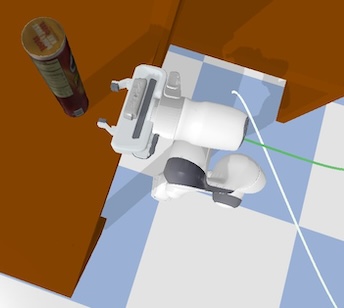}};
\node at (12.75,0) {\includegraphics[height=3.625cm]{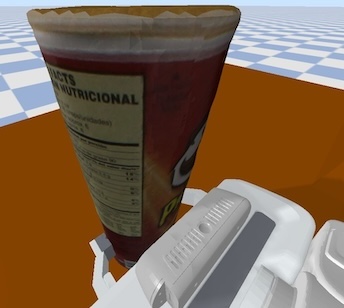}};
\node at (-1.875,1.625) {\textbf{1}};
\node at (2.375,1.625) {\textbf{2}};
\node at (6.625,1.625) {\textbf{3}};
\node at (10.875,1.625) {\textbf{4}};
\end{tikzpicture}
\caption{A sampled trajectory for grasping a can. Frames 1-2, 2-3, 3-4 depict respectively the end-effector alignment with the final pose, the approach stage, and the grasping stage.}
\label{fig:grasping}
\end{figure*}
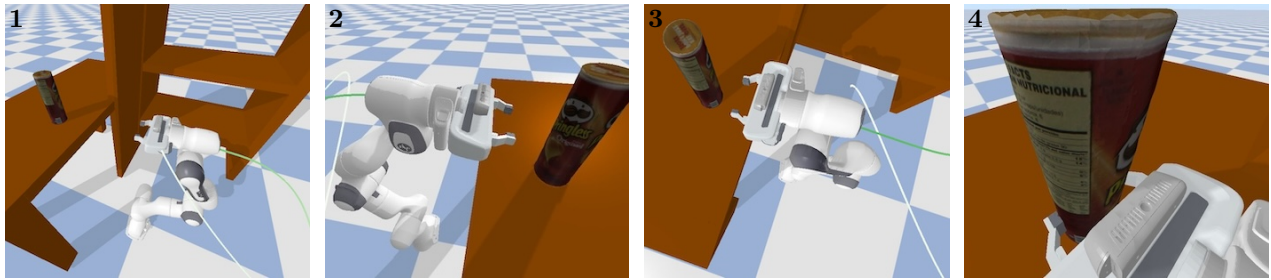

To incorporate constraints that do not have a convenient mathematical form, we include likelihood-based soft constraints in a manner that mirrors non-variational Gaussian Process motion planning algorithms, such as the collision-avoidance term proposed by \textcite{gpmp2}.
In such cases, we replace $\v{f}$ with $\sigma(\v{f})$ to account for inequality constraints as described previously.
This enables us to handle collision avoidance, velocity constraints and other soft constraints, which are implicitly enforced by penalizing violations via the optimization objective, providing a flexible modeling tool that can be adapted to the task at hand.
For example, to enforce grasping, we can include the terms
\[
\label{eq:grasping}
\norm{\f{g}_s - \f{k}_{\f{fwd}}(\sigma(\v{f}))}_{\m\Sigma_{\f{grasp}}}^2
\]
inside the exponential function in the log likelihood \eqref{eq:likelihood}.
Here, $\f{g}_s$ is a \emph{good graspable pose}, which consists of a vector and matrix that, together, represent the position and orientation of the end effector: we subtract element-wise and calculate the norm in the respective product space.
The matrix $\m\Sigma_{\f{grasp}}$ plays a role similar to $\m\Sigma_{\f{obs}}$, and quantifies the relative importance of this term in the likelihood. 
In total, this term effectively allows us to condition the Gaussian process on values in kinematic space, whereas the GP itself lives in joint space.
During the training phase, we draw a batch of samples, calculate the joint's position and orientation for each sample using forward kinematics, and compute the norm.
This likelihood term therefore guides the samples towards the desired end-effector position and rotation, all as part of the variational optimization problem that produces the motion~plan.

As a proof-of-concept, we include a grasping experiment. 
Alignment for viable grasping is achieved by adding an appropriate soft constraint term \eqref{eq:grasping}, which encourages the arm to orient itself accordingly.
One can take this further, conditioning the GP at appropriate time steps to open and close the gripper of the robot to move objects from one place to another.
This reinforces the value of the proposed framework, which allows us to treat different kinds of constraints---here, collision-avoidance and grasping alignment---in essentially-the-same way mathematically.
We demonstrate a motion plan incorporating these elements in \Cref{fig:grasping}, and include a video in the supplementary~material. 

\paragraph{Smoothness}

\begin{figure*}
\begin{subfigure}[b]{0.3\linewidth}
\includegraphics[height=3cm, trim=15pt 20pt 0pt -5pt]{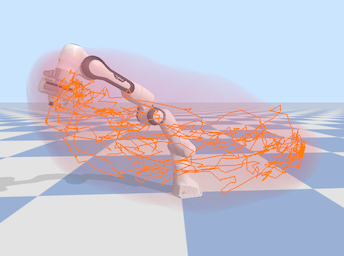}
\vspace*{1ex}
\caption{Matérn-$\sfrac{1}{2}$}
\end{subfigure}
\begin{subfigure}[b]{0.3\linewidth}
\includegraphics[height=3cm, trim=15pt 20pt 0pt -5pt]{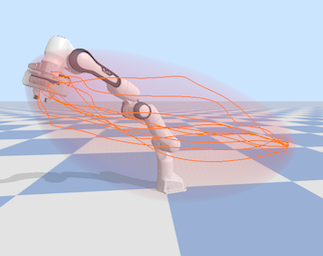}
\vspace*{1ex}
\caption{Matérn-$\sfrac{3}{2}$}
\end{subfigure}
\begin{subfigure}[b]{0.3\linewidth}
\includegraphics[height=3cm, trim=15pt 20pt 0pt -5pt]{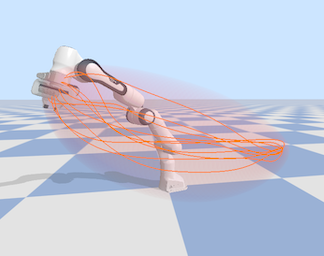}
\vspace*{1ex}
\caption{Matérn-$\sfrac{5}{2}$}
\end{subfigure}
\caption{Illustration of smoothness properties of different kernels, including the nowhere-differentiable Matérn-$\sfrac{1}{2}$, once-differentiable Matérn-$\sfrac{3}{2}$, and twice-differentiable Matérn-$\sfrac{5}{2}$ kernel.}
\label{fig:smoothness}
\end{figure*}

Smoothness can be controlled via kernel choice and its hyperparameters. 
In particular, stationary kernels on $\R^1$ can be formulated to include a \emph{length scale} parameter $\kappa > 0$, and written $k(x,x') = k\del{\frac{|x-x'|}{\kappa}}$.
This parameter determines the characteristic distances over which the function varies \cite{rasmussen-williams}. 
Kernels on $\R^d$ for $d > 1$ can be defined similarly, with one length scale per dimension.
Larger length scales yield smoother trajectories, while smaller length scales yield rougher, more jagged trajectories.
Through kernel choice, for instance by choosing a Matérn kernel with given smoothness \cite{rasmussen-williams}, one can also control properties such as the number of times a motion path is differentiable.
We illustrate this in~\Cref{fig:smoothness}.

\subsection{Computation via Variational Inference and Connections with Other Planners}

By combining the previous likelihood terms, we obtain the negative ELBO
\[
\begin{aligned}
\frac{1}{2} &\E_{q(\v{u})}  \E_{p(\v{f}\given\v{u})} \del[2]{\,\ubr{\norm{\f{h}_\eps(\f{sdf}_s(\f{k}_{\f{fwd}}(\sigma(\v{f}))))}_{\m\Sigma_{\f{obs}}}^2}_{\t{collision term}} 
\\ 
&+ \ubr{\norm{\f{c}(\sigma(\v{f}))}^2_{\m\Sigma_{\f{c}}}}_{\t{soft constraint terms}}\,} + D_{\f{KL}}(q(\v{u}') \from p(\v{u}'\given\v{u}_{\f{c}}))
\end{aligned}
\label{eq:likelihood}
\]
which we minimize to obtain the optimal parameters of $q(\v{u})$.
This enforces equality constraints through the $\v{u}_{\f{c}}$-values, inequality constraints through $\sigma$, and soft constraints through the collision and other soft constraint terms.
Compared to previous GP-based approaches such as for instance \textcite{gpmp2}, we do not need to introduce sparse factor graphs, Laplace approximations, interpolation between time steps, or other scalability tricks.
Instead, we simply carry minimization out using stochastic optimization, by sampling $\v{u}$-values, conditionally sampling $\v{f}$-values, computing the objective, and taking gradient~steps.

From \eqref{eq:likelihood}, we deduce that probabilistic-inference-based motion planners based on variational inference are \emph{stochastic extensions of optimization-based motion planning algorithms}, such as STOMP, CHOMP \cite{stomp, Zucker}, and their variants. 
The main differences in our formulation are: 
\1*[(i)] instead of a single trajectory, we now return a distribution over trajectories, 
\2*[(ii)] the loss contains an extra expectation term, and 
\3*[(iii)] we include an additional KL-based regularizer to ensure we do not stray too far from the prior, allowing us to control smoothness and related task-independent properties in a distributional manner. 
\0*

The effect of these differences is that they introduce uncertainty, at cost of a slightly more complex formulation compared to optimization-based planners.
Our optimization objective is a sum of multiple terms that reflect complementary aspects of motion planning, which need to be weighted appropriately.
Empirically, we found it to be important that the entries of $\m\Sigma_{\f{obs}}$ in \eqref{eq:likelihood} are large enough, to ensure the motion planner does not ignore the collision part of~the~objective.
We present an ablation study in \Cref{app:ablation_section} on how each likelihood term affects the motion planning behavior.
The formulation presented uses a unimodal variational approximation: this choice favors simplicity and ease of optimization.
One could instead make a choice in favor of the opposite tradeoffs, following for instance \textcite{ewerton}, by using a mixture model for the variational approximation.

Compared to other probabilistic-inference-based planners, such as GPMP2 \cite{gpmp2}, our formulation ensures computational tractability through sparse GPs, waypoints, and variational inference, rather than through a factor graph arising from the SDE prior.
Compared to GVI \cite{gvi}, we work with posterior function samples rather than purely in distribution-space, making our implementation much closer to that of an optimization-based planner.
Our algorithm's complexity is $\c{O}(TM^2 + dM^3 + T(d + s))$ for obtaining the variational parameters which define a motion plan, $\c{O}(dM^3)$ for computing error bars, and $\c{O}(TM^2 + M^3)$ for both computing the mean motion plan and drawing one Monte Carlo sample when using efficient sampling via pathwise conditioning \cite{wilson,wilson_jmlr}, and $\c{O}(T(d+s))$ to compute the likelihood. Here, $T$ is the number of time points, $M$ is the number of inducing points, which for our tasks is usually around a dozen, $d$ is the dimensionality of joint space, and $s$ is the number of spheres.
Calculations are given in \Cref{app:time complexity}.

\paragraph{Replanning}
In situations where we need to replan, we do so by starting with the previous motion plan and moving inducing points and inducing values around appropriately in order to compute a new motion plan. 
We also condition on the current value as initial state.

\subsection{Uncertainty: Intervals and Monte Carlo Function Samples}

Once a Gaussian process motion planning algorithm has completed the optimization process, the next step is to extract an actual motion plan from it.
Since, in the variational framework, motion plans map bijectively to Gaussian processes, the \emph{most likely} motion plan under the variational posterior can be obtained via the variational posterior's mean .
We can incorporate uncertainty in two ways: (a) through posterior intervals or (b) by drawing Monte Carlo random function samples.
These work in the following manner:

\1*[(a)] Posterior intervals: for a given scaling level $\alpha > 0$ and independent GPs per output dimension, we can represent uncertainty through posterior intervals, which are $\mu_{f\given\v{u}}(x) \pm \alpha\sqrt{k_{f\given\v{u}(x,x)}}$. For large-enough $\alpha$, this gives a high-probability estimate on the region through which motion plans will travel.
\2*[(b)] Posterior function samples: we can also draw random function $\phi \~[GP](\mu_{f\given\v{u}}, k_{f\given\v{u}})$. 
For stationary kernels, we do so up to a small error tolerance using the \emph{efficient sampling} technique of \textcite{wilson,wilson_jmlr}, which works by drawing approximate random functions from the prior, and transforming them into the posterior using pathwise conditioning. 
Once drawn, we can then plug these randomly-sampled functions into upstream systems as needed for the given task.
\0*

\begin{table*}[t]
\footnotesize
\begin{tabular}{lllcccccc}
\toprule
\multirow{2.5}{*}{\textbf{Robot}} & \multirow{2.5}{*}{\textbf{Env}} & \multirow{2.5}{*}{\textbf{M}} & \multicolumn{4}{c}{\textbf{Planners}} \\
\cmidrule{4-8}
 & & & {vGPMP (ours)} & {RRT-connect} & {LBKPIECE} & {CHOMP}  \\
\midrule
\multirow{6}{*}{Franka} & \multirow{3}{*}{I} & Acc & $\v{100\%}$ & $\v{100\%}$ & $88.89\%$ & $18.06\%$  \\
     & & MPL & $2.75 \pm 2.14$& $1.16 \pm 1.67$ & $1.82 \pm 2.48$ & $0.96 \pm 1.95$ \\ 
     & & MC & $\v{-1.17 \pm 1.73}$ & $-11.81 \pm 3.23$ & $-14.03 \pm 4.55$ & $-9.25 \pm 6.94$  \\
\cmidrule{2-8}
& \multirow{3}{*}{B}  & Acc & $81.81\%$ & $\v{100\%}$ & $\v{100\%}$ & $14.54\%$ \\ 

     & & MPL & $2.60 \pm 1.86$ & $2.10 \pm 2.09$ & $2.21 \pm 2.19$ & $1.27 \pm 2.01$  \\ 
     & & MC & $\v{-6.76 \pm 4.69}$ & $-84.42 \pm 18.63$ & $-116.58 \pm 20.05$ & $-37.47 \pm 12.77$  \\
\midrule
\multirow{6}{*}{UR10} & \multirow{3}{*}{I} & Acc &$ \v{100\%}$ & $97.22\%$ & $86.11\%$ & $11.11\%$  \\ 
     & & MPL & $1.63 \pm 1.09$& $2.42 \pm 2.55$ & $4.03 \pm 4.37$ & $0.20 \pm 0.47$  \\ 
     & & MC & $\v{-56.03 \pm 11.86}$ & $-281.84 \pm 22.91$ & $-479.73 \pm 35.52$ & $-5940.77 \pm 108.31$ \\
\cmidrule{2-8}
& \multirow{3}{*}{B} & Acc & $\v{89.82\%}$ & $85.45\%$ & $16.36\%$ & $18.18\%$ &\\ 
     & & MPL & $3.95 \pm 1.91$ & $1.82 \pm 2.22$ & $0.93 \pm 1.56$ & $0.55 \pm 0.90$  \\ 
     & & MC & $\v{-7.10 \pm 5.55}$ & $-42.56 \pm 13.03$ & $-26.48 \pm 9.97$ & $-58.53 \pm 6.58$  \\
\bottomrule
\end{tabular}
\caption{Experiments in different environments using two robots. The environments (Env) used are named Industrial (I) and Bookshelves (B), and the Measures (M) are accuracy (Acc), Mean Path Length (MPL), and Mean Clearance (MC). Mean and standard deviation are computed over the entire benchmark suite, and are averaged over five total runs.}
\label{tbl:non_gp_comparison}
\end{table*}

We emphasize that (b) is computationally tractable in part because the variational formulation can handle general kernels.
These technical primitives can be used to propagate uncertainty into any upstream decision-making algorithm built atop of GP.
Gaussian processes are widely used to represent uncertainty in a number of decision-making systems, including in Bayesian optimization, active learning, and reinforcement learning.
Our formalism can therefore in principle facilitate the development of similar systems in the context of motion planning with further work.

\section{Experiments}
\label{sec:experiments}

We now evaluate vGPMP and compare its performance and capabilities with other motion planners, including both those based on probabilistic inference and those based on other formalisms.
To illustrate the unifying nature of our techniques, we consider standard collision-avoidance benchmarks.
In all cases, full experiment details can be found in \Cref{app:experiments}.

To understand how vGPMP and other motion planners handle collision avoidance, we ran them on two kinds of motion planning benchmarks.
For the first kind, we compare with a state-of-the-art probabilistic-inference-based planner \emph{GPMP2}, using similar hyperparameters as detailed in \Cref{app:experiments}, on the Barrett \emph{WAM} robot (7 DOF) in the \emph{Lab} environment \cite{gpmp}.
For the second kind, we compare with sampling-based and optimization-based planners, using benchmarks originally introduced by \textcite{schulman2013}, which are based on the \emph{Franka Panda} (7 DOF) and \emph{UR10} (6 DOF) robots in the \emph{Industrial} and \emph{Bookshelves} environments, and have 36 and 55 unique motion planning problems, respectively.
A visualization of these environments can be seen in \Cref{app:experiments}.

We evaluate behavior with respect to (i) \emph{accuracy}, which is the percentage of tasks within the benchmark suite correctly solved by the planner, (ii) \emph{clearance}, which quantifies collision-avoidance by measuring the distance to objects, and (iii) \emph{path length}, which is the total distance moved by the end effector. 
Specifically, clearance, which we use it as a metric to assess safety, is defined as the negation of the collision term $-\frac{1}{2}\norm{\f{h}_\eps(\f{sdf}_s(\f{k}_{\f{fwd}}(\sigma(\v{f}))))}_{\m\Sigma_{\f{obs}}}^2$ used in the likelihood.
Note that clearance and path length are related: higher clearance pushes paths further away from objects, hence increasing path length. 
If required, an extra objective can be added to bring paths to desired lengths whilst maintaining collision avoidance.

All measurements are repeated using 5 different random seeds to assess variability.
Further details on measurements are given in \Cref{app:experiments}.

\Cref{tbl:gpmp2_comparison} shows that vGPMP and GPMP2 both succeed in finding obstacle-free paths reaching the targets in all cases.
Compared to GPMP2, vGPMP leads to better clearance values and consequently longer paths moving away from the objects ensuring safety. 
Both methods require careful tuning of $\m\Sigma_{\f{obs}}$ to ensure that this term is a sufficiently-important part of the loss to achieve collision-avoidance. Overall, in comparison to GPMP2, vGPMP performs comparably and provides a simpler computational pipeline and more general formulation.

\begin{figure}[t!]
\begin{subfigure}[b]{0.49\linewidth}
\includegraphics[height=3.5cm]{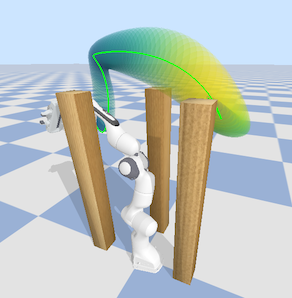}
\caption{Start query state}
\label{fig:boxes_start}
\end{subfigure}
\begin{subfigure}[b]{0.49\linewidth}
\includegraphics[height=3.5cm]{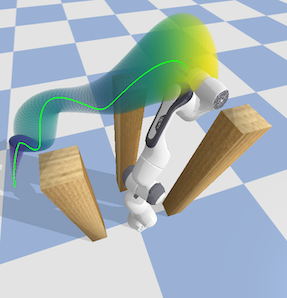}
\caption{Goal query state}
\label{fig:boxes_end}
\end{subfigure}
\caption{Example distribution of collision-free paths generated by vGPMP in the boxes environment, with a random sample shown in green.}
\label{fig:dist_env}
\end{figure}

Next, we examine how uncertainty translates to real-world execution.
For vGPMP, we use Monte Carlo sampling to obtain 150 paths, from which we calculate the resulting envelope around the mean trajectory.
The intervals in \Cref{fig:dist_env} show the successful incorporation of the problem constraints as expressed in the low uncertainty at the beginning and end of the trajectory.  
This uncertainty grows as one moves further away from start and goal states, as well as points of collision.

\begin{figure*}[b!]
\begin{tikzpicture}
\node at (0,0) {\includegraphics[height=2.75cm]{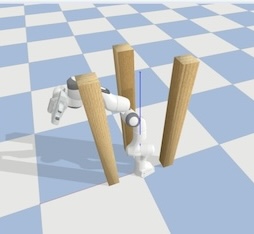}};
\node at (3.25,0) {\includegraphics[height=2.75cm]{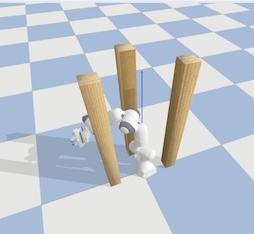}};
\node at (6.5,0) {\includegraphics[height=2.75cm]{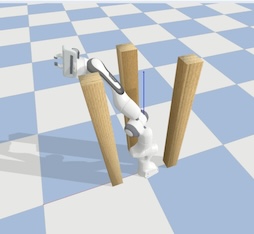}};
\node at (9.75,0) {\includegraphics[height=2.75cm]{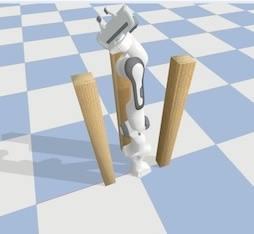}};
\node at (13,0) {\includegraphics[height=2.75cm]{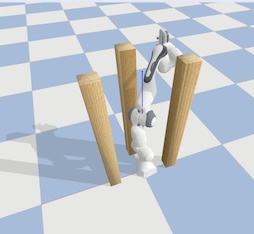}};
\node at (0,-3) {\includegraphics[height=2.75cm]{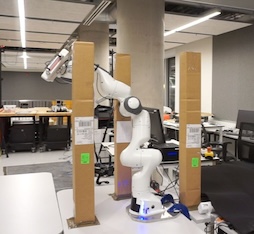}};
\node at (3.25,-3) {\includegraphics[height=2.75cm]{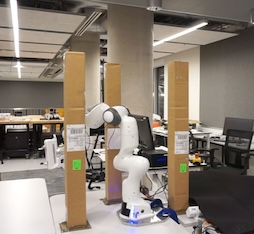}};
\node at (6.5,-3) {\includegraphics[height=2.75cm]{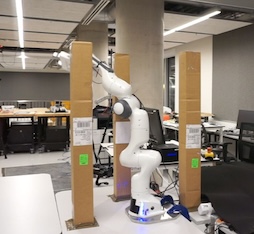}};
\node at (9.75,-3) {\includegraphics[height=2.75cm]{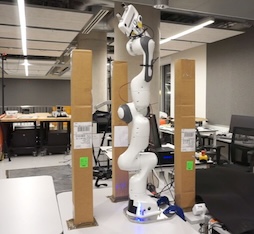}};
\node at (13,-3) {\includegraphics[height=2.75cm]{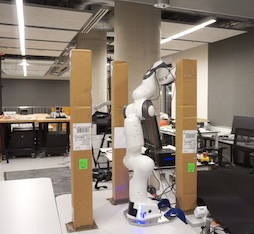}};
\end{tikzpicture}
\caption{Sampled trajectory executions on the real (bottom) and the simulated (top) robot showing successful transfer using a low-level controller.}
\label{fig:panda-seq}
\end{figure*}

\Cref{tbl:non_gp_comparison} shows that vGPMP consistently leads to high success rates in reaching the goal states, achieving 91.42\% across all experiments, which is significantly higher than the performance of LBKPIECE (69.78\%) and CHOMP (15.69\%). 
RRT-connect achieves a slightly higher success rate, 95.05\% on average---however, the generated paths demonstrate worse clearance values.
In most cases, vGPMP produces longer paths, since uncertainty drives paths away from obstacles for better collision avoidance. 

\begin{table}[t!]
\begin{tabular}{c c c}
\toprule
\textbf{Metric}   & \textbf{vGPMP (ours)} &  \textbf{GPMP2} \\ \midrule
Acc.   & $\v{100\%}$ &   $\v{100\%}$   \\ 
MC &  $\v{-0.90 \pm 1.50}$   & $-2.58 \pm  2.54$ \\ 
MPL & $2.17 \pm 1.00$ &  $1.89 \pm 0.98$ \\ 
\bottomrule
\end{tabular}
\caption{Comparison using WAM robot in the lab environment. Metrics shown are \emph{accuracy}, \emph{mean clearance}, and \emph{mean path length}, defined in \Cref{sec:experiments}.}
\label{tbl:gpmp2_comparison}
\end{table}

vGPMP consistently yields better clearance values in all environments using the two robots, except with \emph{Franka Panda} in the \emph{bookshelves} environment, where the performance of vGPMP drops due to collisions with the floor: this could be alleviated by including an extra term in the loss, which we omitted to ensure a fair comparison with other planners that also do not consider the floor. 
In total, without post-processing, the uncertainty produced by vGPMP results in safer trajectories compared to those of other planners.

Finally, we apply our approach directly to a real robot. 
We demonstrate the feasibility of the approach by executing a trajectory generated from vGPMP given three obstacles placed around a Franka Panda robot. 
To execute the generated motion plans on the physical robot, we used a low-level joint impedance position controller \cite{franzese2021ilosa}.
We show the resulting trajectory in \Cref{fig:panda-seq}, and provide a video in the paper's supplementary materials.

\section{Conclusion}

We present a unifying framework for Gaussian-process-based motion planning with obstacle avoidance, using a formulation based on variational inference.
The framework supports equality constraints, inequality constraints, and general soft constraints. It also allows for fine-grained control over smoothness through the prior.
Our approach generalizes previous Gaussian-process-based motion planners to arbitrary kernels in a manner that simplifies computation.
Additionally, by viewing variational inference as an optimization problem over a space of random motion paths, our framework also generalizes optimization-based planners to incorporate uncertainty.
We evaluate the approach using different robots, environments, and tasks---including obstacle avoidance and grasping---showing that it has high accuracy in reaching target positions, while avoiding obstacles and using uncertainty to providing better clearance compared to baselines. We demonstrate the feasibility of the approach on a real robot, executing a sampled trajectory while avoiding obstacles.

\printbibliography

\appendix

\onecolumn
\toptitlebar 
{\centering{\Large\bfseries Supplementary Materials:\\A Unifying Variational Framework for \\ Gaussian Process Motion Planning\par}}
\bottomtitlebar
 
\section{Algorithm}
\label{app:vgpmp_algorithm}

A full algorithmic description of vGPMP is given below.

\begin{algorithm}
\caption{vGPMP}\label{pseudocode}
\begin{algorithmic}[1]
\State \textbf{Inputs}: number of total optimization steps $L$, the variational distribution $q(\v{u})$, the prior $p(\v{u})$, input timesteps $\m{X} = \left[t_0, t_1, \cdots, t_N \right]$, the process $f \sim \c{GP}(\mu, k)$, step size $\rho$, total number of samples $K$ drawn at each ELBO estimation step, samples drawn from posterior $P$, optional: query timesteps with finer discretization e.g. $\m{X}_{\f{new}} = \left[t_0, t_{0.5}, \cdots, t_{N_{\f{new}}} \right]$, $N_{\f{new}}>N$.
\Statex
\For{$\ell = 1:  L$}
    \State \emph{Draw }$\left\{\v{u}^k\right\}_{k=1}^K \sim \c{N}(\v{\mu}_{\v{u}_\ell}, \m{\Sigma}_{\v{u}_\ell})$
    \State $\m{K}_{\v{z}\v{z}} \gets k(\v{u}, \v{u})$
    \State $\mathbf{L} \gets \f{CholeskyDecomposition}(\m{K}_{\v{zz}})$ 
    \LeftComment{Matrix inversion using Cholesky decomposition}
    \State $\m{K}_{\v{zz}}^{-1} \gets \f{CholeskySolve}(\m{L}, \m{I})$ 
    \LeftComment{Draw pathwise samples}
    \State $\left\{\v{f}^k\right\}_{k=1}^K \sim f(\m{X}) + \m{K}_{\v{xz}}\m{K}_{\v{zz}}^{-1}(\v{u}-\v{f}_m)$

    \LeftComment{Evaluate the likelihood and average over samples}
    \State $\mathbf{E} \gets \sum_{i=1}^K \,\ubr{\norm{\f{h}_\eps(\f{sdf}_s(\f{k}_{\f{fwd}}(\sigma(\v{f}^i))))}_{\m\Sigma_{\f{obs}}}^2}_{\t{collision term}} + \ubr{\norm{\f{c}(\sigma(\v{f}^i))}^2_{\m\Sigma_{\f{c}}}}_{\t{soft constraint terms}}\,$    

    \LeftComment{Compute the mean and covariances for $p(\v{u})$ and $q(\v{u})$}
    \State $\v{\mu_p} \gets \m{K}_{\v{zz}}^{-1} \v{u}_c$
    \State $\v{\mu_{q}} \gets [\v{u}_c, \v{\mu}_{\v{u}_s}]$

    \LeftComment{Whiten the difference between $q(\v{u})$ and $p(\v{u})$}
    \State $\m{W} \gets \f{TriangularSolve}(\m{L}, \v{\mu_q} - \v{\mu_p})$

    \LeftComment{Compute the KL divergence and ELBO}
    \State $\f{D}_{\f{KL}} \gets 0.5 \left(\log|\m{\Sigma}_{\v{u}_\ell}| - \tr(\m{\Sigma}_{\v{u}_\ell}) - \dim(\m{W}) + \m{W}^\top \m{W}\right)$
    \State $\c{L} \gets \alpha \frac{1}{N}\sum_{i=1}^N \m{E}^i + \f{D}_{\f{KL}} $
    \LeftComment{Update the variational distribution}
    \State $\v{\mu}_{\v{u}_{\ell+1}}=\v{\mu}_{\v{u}_\ell} + \rho \left[\nabla_q\c{L}\right]$
    \State $\m{\Sigma}_{\v{u}_{\ell+1}}=\m{\Sigma}_{\v{u}_\ell} + \rho \left[\nabla_q\mathcal{L}\right]$
\EndFor
    \State \emph{Draw }$\left\{\v{u}^i\right\}_{i=1}^P \sim \c{N}(\v{\mu}_{\v{u}_{L}}, \m{\Sigma}_{\v{u}_{L}})$
    \LeftComment{Draw pathwise samples from posterior or return the mean}
    \State $\left\{\v{f}^i\right\}_{i=1}^P \sim f(\m{X}_{\f{new}}) + \m{K}_{\v{x'z}}\m{K}_{\v{zz}}^{-1}(\v{u}-\v{f}_m)$
    \LeftComment{Optional: pick a trajectory of choice - here we pick the lowest collision cost trajectory}
    \State $\v{\theta} = \argmin_{\left\{\v{f}^i\right\}_{i=1}^P}\,\left[\norm{\f{h}_\eps(\f{sdf}_s(\f{k}_{\f{fwd}}(\sigma(\v{f}^i))))}_{\m\Sigma_{\f{obs}}}^2\,\right]$  
    \State \textbf{return:} $\v{\theta}$
\end{algorithmic}
\end{algorithm}

\section{Incorporating Additional Constraints under the Variational Representation}
\label{app:constraints}

For learning the variational distribution, we choose $q(\v{u}) = \c{N}(\v\mu_{\v{u}}, \m\Sigma_{\v{u}})$ with a particular representation we now describe.
Letting $\v{u} = [\v{u}',\v{u}_{\f{c}}]$ be the partition of $\v{u}$ into the learned and constrained parts, the matrix $\m\Sigma_{\v{u}}$ contains one part for $\v{u}'$, which is padded with zeros to encode the uncertainty in the given motion planning constraints $\v{u}_{\f{c}}$. 
One can additionally introduce noise in the latent variables, but we omit this here to ease notation.

\paragraph{Computational Considerations} 
To improve computational efficiency, we use a whitened representation that only considers the learned portion of the covariance $\m\Sigma_{\v{u}}$. 
Let $\v{z}'$ and $\v{z}_{\f{c}}$ be the inducing points corresponding to $\v{u}'$ and $\v{u}_{\f{c}}$, respectively.
The part of the distribution representing $\v{u}_{\f{c}}$ is a degenerate Gaussian, so we omit it from the whitening process. 
This representation is achieved using the block partitioned representation of $\m{K}_{\v{z}\v{z}}$, because the Cholesky decomposition of the covariance of $p(\v{u}' | \v{u}_{\f{c}})$ can be written as $\f{Chol}(\m{K}_{\v{z}'\v{z}'} - \m{K}_{\v{z}'\v{z}_{\f{c}}} \m{K}_{\v{z}_{\f{c}} \v{z}_{\f{c}}}^{-1} \m{K}_{\v{z}_{\f{c}} \v{z}'})$: this can be seen by considering the block LU decomposition. 
An added benefit of this approach is that the KL term in \eqref{eq:likelihood} can be computed efficiently, as the Cholesky factor is already cached and used in the pathwise updates.

\paragraph{Velocity Modeling and Higher Order Derivatives}

Including constraints on derivatives, such as maintaining constant velocity, is crucial for real-world execution.
By incorporating these constraints, the resulting models can accurately account for physical system limitations, ensuring the feasibility and reliability of planned trajectories during execution. 
In our framework, one can encode velocities as prior information by considering inducing variables along with their time derivatives, namely $\v{u} = [\v{u}', \dot{\v{u}}', \v{u}_{\f{c}}, \dot{\v{u}}_{\f{c}}]$ which implies that the respective kernel matrix can be assembled by differentiating the kernel function. 
In many situations, we set the start and end velocities to zero. 
Our framework also allows angular velocities of the samples to be computed by differentiating the sample paths.
This means they can be included in the likelihood via a soft constraint term which brings the velocity of sampled trajectories to desired values.
This lead to, for instance, the likelihood term
\[
\norm[1]{\dot{\sigma}(\v{f}) - \mu_\m{v}}_{\m\Sigma_{\f{velocity}}}
.
\]
Acceleration and other higher-order derivatives can also be considered following a similar argument, assuming appropriate kernel choice to ensure sufficient differentiability. 
For the Matérn-$\sfrac{5}{2}$ kernel, this benefit is two-fold as the GP is then Markovian: if advantageous in the setting at hand, this can be used to re-formulate the necessary computations in terms of sparse linear algebra, which makes it possible to generalize the stochastic-differential-equation-based techniques of \textcite{gpmp2} to GP priors with more available derivatives.

\section{Time Complexity}
\label{app:time complexity}

In this section, we examine the scalability of vGPMP.
Let $s$ be the number of spheres that model the robot body. 
Let $T$ be the total number of discrete time points.
Let $M$ be the number of inducing points, and $d$ be the dimensionality of joint space.
The time complexity of \emph{one iteration} can be split up into
\[
\text{Cost}_\text{iteration} = C_\text{sampling} + C_\text{likelihood} + C_\text{KL}.
\]
We begin by deriving the complexity cost of computing the likelihood for a single sample, for one single joint configuration at some timestep $t$.

The time complexity of the forward kinematics computation using the Denavit-Hartenberg (DH) convention and homogeneous transformation matrices is $\c{O}(d)$, where $d$ is the number of degrees of freedom of the robot arm. To compute the forward kinematics, we need to multiply the homogeneous transformation matrices for each joint, as follows:
\[
\m{F}_d = \prod_{i=0}^{d-1} \m{F}_{i,i} \m{F}_{i,i+1}
\]
Each $\m{F}_i$ is $4\x4$ matrix computed using the DH parameters, and hence requires constant amount of computation.
Then, an additional transform is needed to translate the sphere offsets into world coordinates. 
This can be easily achieved by considering some other homogeneous transform $\m{H}_{i,j}$ for some sphere $c_j, j=\{1, \.\.\., s\}$ and some frame of reference (in world coordinates) $i$ such that we can get the sphere placement in world coordinates, namely
\[
\m{H}_j = \m{F}_i \m{H}_{i,j}
.
\]

\begin{table}[t]
\footnotesize
\begin{tabular}{llcccccc}
\toprule
\multirow{2}{*}{\textbf{Robot}} & \multirow{2}{*}{\textbf{Env}} & \multicolumn{5}{c}{\textbf{Parameters}} \\
\cmidrule{3-7}
& & $\E[q(\v{u})]$ & length scale & kernel variance & $\sigma_{\f{obs}}$  & Ind. var. \\
\midrule
\multirow{4}{*}{Franka} & \multirow{2}{*}{I} & Zeros & $[2, 2, 2, 2, 2, 2, 2]$ & 0.2 & 0.005 & 10 \\
& & Trained & Trained & Trained & Not trained & Trained \\ 
\cmidrule{2-7}
& \multirow{2}{*}{B}  & Interp. & $[4, 4, 4, 3, 3, 3, 3]$ & 0.5 & 0.0005 & 24 \\ 
& & Trained & Trained & Trained & Not trained & Trained \\ 
\midrule
\multirow{4}{*}{UR10} & \multirow{2}{*}{I} & Interp. &$[6, 6, 6, 6, 6, 6]$ & 0.1 & 0.0001 & 18 \\ 
& & Trained & Trained & Trained & Not trained & Trained \\ 
\cmidrule{2-7}
& \multirow{2}{*}{B} & Upright & $[4, 4, 4, 4, 4, 4]$ & 0.25 & 0.0005 & 12 \\ 
& & Trained & Trained & Trained & Not trained & Not trained \\ 
\midrule
\multirow{2}{*}{WAM}  
& \multirow{2}{*}{Lab}  & Interp. & $[5, 5, 5, 4, 5, 5, 5]$ & 0.05 & 0.005 & 14 \\ 
& & Trained & Trained & Trained & Not trained & Not trained \\ 
\bottomrule \\
\end{tabular}
\caption{Experiment configurations for various robots in different environments. The environments (Env) used are named \emph{industrial} (I) and \emph{bookshelves} (B). Interp. stands for \emph{interpolated}. In all cases, set $\eps = 0.05$, aside from UR10 on the \emph{industrial} environment where it was $\eps = 0.08$ and WAM on the \emph{lab} environment where it was $\eps = 0.03$. We also list which parameters are optimized during training, which are denominated as \emph{trained}, and which are fixed during training, denominated as \emph{not trained}.}
\label{tab:hyperparameters}
\end{table}

To do this for all joints and all spheres, the time complexity is $\c{O}(s)$, which brings the overall cost to $\c{O}(d+s)$. 
After the joint configuration has given the robot pose and the sphere position in world coordinates, the likelihood can be computed. 
The cost can be broken up into applying the hinge loss to the signed distances, and the cost of computing the quadratic form.
For the first half, the cost of computing the signed distance involves a lookup in a 3D grid, incurring constant cost per sphere.
Since the hinge loss' complexity is linear in the number of spheres, the overall time complexity of this procedure is $\c{O}(s)$.
Additionally, since the covariance-matrix-term in the likelihood is assumed diagonal, for a given 
vector of spheres of size $s$, the quadratic form can be computed in linear time $\c{O}(s)$.
At each iteration of the optimization the likelihood is evaluated at $T$ timesteps, bringing the overall cost to $C_{\text{likelihood}} = T(d+s)$.

The efficient sampling technique of \textcite{wilson,wilson_jmlr} uses pathwise conditioning, which involves solving a linear system and computing the Cholesky decomposition of $\m{K}_{\v{z}\v{z}}$.
Therefore, drawing the samples amounts to $C_{\text{sampling}} = TM^2 + M^3$ for one sample. We view the cost of Fourier features as~constant.  

Lastly, we consider computation of the KL term.
The dominant factor here is the computation of the covariance matrix inverse and determinant.
We must also whiten $\m\Sigma_{\v{u}}$. We have that (a) the cost of computing $\m{K}_{\v{z}\v{z}}$ is $\c{O}(dM^2)$, (b) the cost of computing the Cholesky decomposition and solving the linear system takes $\c{O}(dM^3)$ time for a dense matrix, (c) the cost of whitening the covariance is dominated by solving the linear system, where the Cholesky factor is computed beforehand and is upper triangular, which therefore also takes $\c{O}(dM^3)$. 
Overall, $C_\text{KL}=dM^3$, which gives the overall time complexity of \Cref{pseudocode} as $\c{O}(TM^2 + dM^3 + T(d + s))$.

\section{Experimental Details}
\label{app:experiments}

Experiments were performed on an AMD Ryzen 7 3700x CPU. For the proposed planner, we employ independent GPs per output dimension with Matérn-$\sfrac{5}{2}$ kernels. This GP model provides a sufficient degree of differentiability to express higher order dynamics, and is expressive enough to accommodate different ranges of motion necessary in the motion plans. 
Experiments were repeated five times with different seeds.
We use the GPFlow library \cite{gpflow}, with the complementary GPFlow-based pathwise sampling module \cite{wilson,wilson_jmlr} and PyBullet for simulations \cite{pybullet}. 
vGPMP uses Adam with hyperparameters $\beta_1=0.8$ and $\beta_2=0.95$ and learning rate $\eta=0.09$. 

Since optimization is carried out on the loss directly, any standard optimizer can be used.
In practice, with standard stochastic optimization we observed convergence across all test suites in 150 optimization steps or less---however, if even faster convergence is desired, one could for instance apply natural gradients with some additional computational overhead \cite{khan-nielsen,salimbeni2018natural}.

\begin{figure}[t]
\begin{subfigure}{0.49\textwidth}
\includegraphics[height=4.5cm]{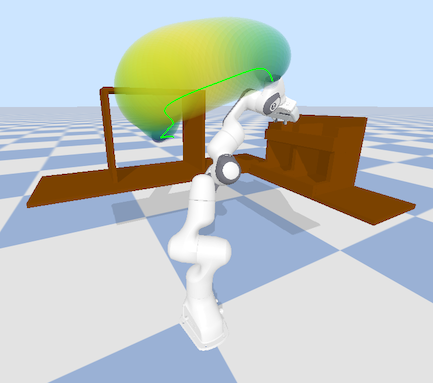}
\caption{Industrial environment}
\label{fig:industrial}
\end{subfigure}
\begin{subfigure}{0.49\textwidth}
\includegraphics[height=4.5cm]{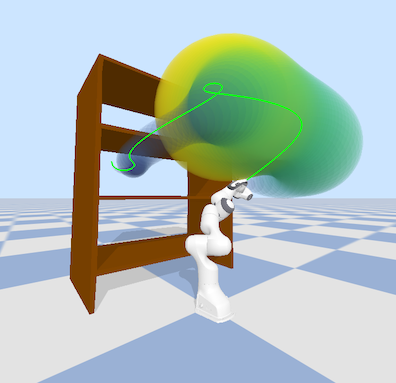}
\caption{Bookshelves environment}
\label{fig:bookshelves}
\end{subfigure}
\caption{Example distributions of collision-free paths generated by vGPMP in two benchmark environments. Sampled trajectories are shown in green. }
\label{fig:envs}
\end{figure}

We include RRT-connect, LBKPIECE and CHOMP via the MoveIt! OMPL library \cite{moveit}. 
RRT-connect and LBKPIECE were ran with default settings, with post-processing left on to smoothen the resulting trajectory. 
CHOMP was ran using the configuration detailed by \textcite{gpmp2}. 

We compare the proposed approach, vGPMP, with GPMP2 \cite{gpmp2} in the \emph{lab} environment, originally introduced by \textcite{gpmp}, using the (7 degree of freedom) Barrett WAM robot. 
We test both planners using 24 different cases with various start and goal states.
We initialize GPMP2 with parameters as outlined in \textcite{gpmp2}: $\m\Sigma_{\f{obs}}=0.02^2\m{I}$, $\eps = 0.2$ with 101 temporally equidistant states. 
Results are detailed in \Cref{tbl:gpmp2_comparison}. 
For vGPMP, we set $\m\Sigma_{\f{obs}}=0.005^2\m{I}$, $ \eps=0.03$, and only 14 inducing locations. For optimization, we use 14 samples per iteration for a total of 150 iterations. 

All parameters used for baselines against RRT-connect, LBKPIECE, CHOMP and GPMP2 are outlined in \Cref{tab:hyperparameters}. 
The number of iterations for \emph{Franka} and \emph{UR10} is 130. 
From \Cref{tbl:non_gp_comparison}, in the \emph{industrial} environment using UR10 our approach finds shorter paths in general, which is in part due to higher length scale initialization. 

\begin{figure}[b]
\includegraphics[height=4.5cm]{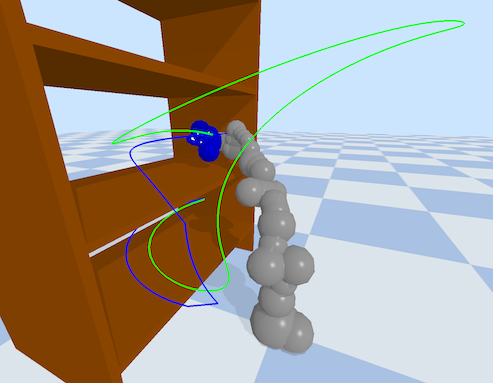}
\caption{Comparison of trajectories from a baseline motion planner, RRT-connect (in blue) and vGPMP (in green). As in this example, in general vGPMP leads to smooth trajectories further away from the obstacles. Spheres around robot links that are used in collision detection can also be seen.}
\label{fig:rrt_vs_vgpmp}
\end{figure}

We evaluated the performance of vGPMP and other baseline planners in two different environments, which can be seen in Figure~\ref{fig:envs}, using various start and goal states. 
vGPMP generates smooth trajectories that are also further away from the obstacles in comparison to baselines as illustrated in \Cref{fig:rrt_vs_vgpmp}. If we filter samples to obtain a \emph{most-collision-free path}, this avoids the bookshelf in a safer manner than the respective path from RRT-connect. 
To ensure balanced comparisons, average success rates reported are calculated as a weighted average with respect to the number of problems.

Each hyperparameter in \Cref{tab:hyperparameters} can be interpreted as follows. The \emph{expectation} $\E[q(\v{u})]$ represents the mean of the prior around which paths are sampled for optimization.
\emph{Length scale} refers to the kernel length scale hyperparameter, which controls spatial variability---see \textcite{rasmussen-williams}. \emph{Kernel variance} corresponds to the size of the outlined volume in \Cref{fig:dist_env}. The \emph{likelihood standard deviation} term $\sigma_{\f{obs}}$ controls the weight on the likelihood in the objective, with lower values representing higher weight on the likelihood. The term \emph{Ind. Var.} represents the number of inducing variables corresponding to waypoints used during optimization.

\begin{figure}[t]
  \begin{subfigure}{0.49\textwidth}
    \centering
    \includegraphics[height=4.375cm]{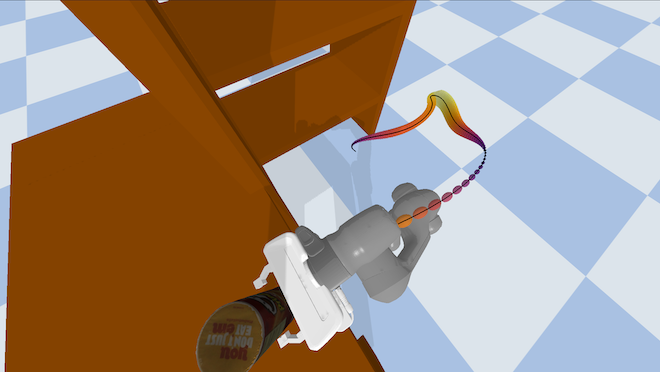}
    \caption{Trajectory distribution for the Panda arm using all likelihood terms.}\label{fig:graspingexample_successful}

  \end{subfigure}\hfill
  \begin{subfigure}{0.49\textwidth}
    \centering
    \includegraphics[height=4.375cm]{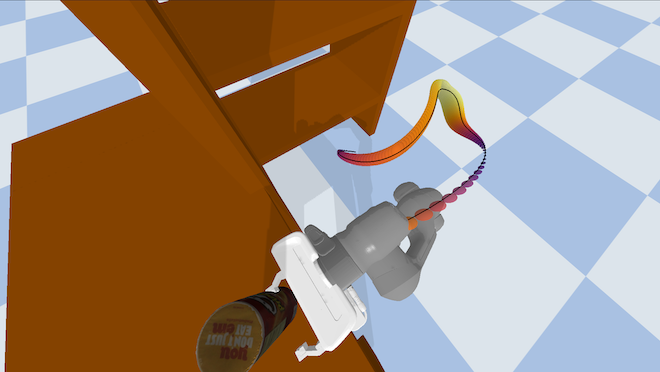}
    \caption{Trajectory distribution for the Panda arm missing the collision likelihood.}\label{fig:graspingexample_coll}
  \end{subfigure}
  
  \vspace{0.5\baselineskip} % Adjust vertical space between rows
  \begin{subfigure}{0.49\textwidth}
    \centering
    \includegraphics[height=4.375cm]{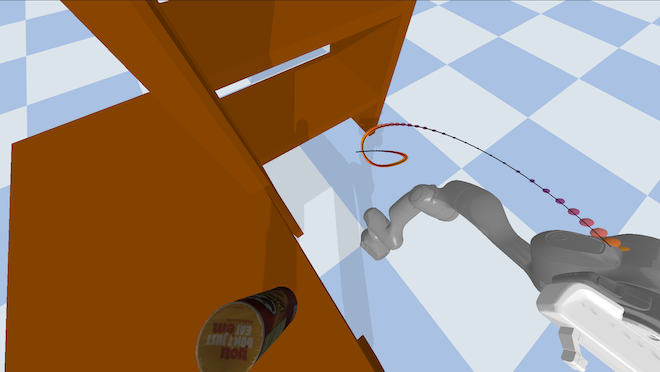}
    \caption{Trajectory distribution for the Panda arm missing the position likelihood. }\label{fig:graspingexample_pos}
  \end{subfigure}\hfill
  \begin{subfigure}{0.49\textwidth}
    \centering
    \includegraphics[height=4.375cm]{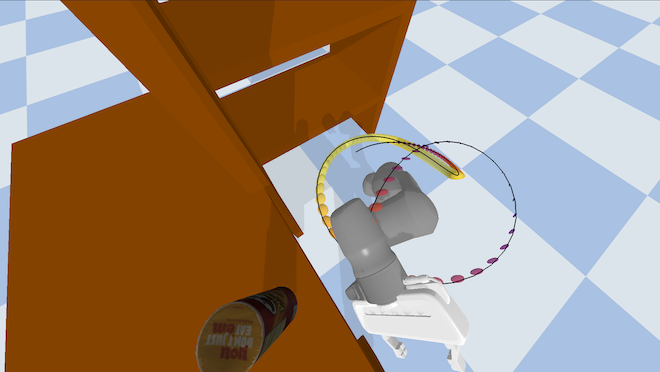}
    \caption{Trajectory distribution for the Panda arm missing the orientation term.}\label{fig:graspingexample_or}
  \end{subfigure}
  \caption{Visual comparison of the influence of each constraint term on optimization and the resulting trajectory distribution. In \Cref{fig:graspingexample_successful} optimization uses all three likelihoods for optimization: collision, position and orientation. In \Cref{fig:graspingexample_coll} the collision likelihood was omitted. In \Cref{fig:graspingexample_pos}, \Cref{fig:graspingexample_or} the position and orientation part of the grasping likelihood were omitted, respectively. }\label{fig:grasping_experiment_showcase}
\end{figure}

\section{Ablation Study}\label{app:ablation_section}

In \Cref{sec:likelihoods}, we showed that our framework can be easily adapted to include additional constraints of interest. 
Here, we use the grasping setup using the additional terms introduced in \eqref{eq:grasping} to perform an ablation experiment on how the contribution of each likelihood term affects the final distribution.
For all experiments, the configuration of vGPMP is fixed. 
\Cref{fig:graspingexample_successful} shows the final trajectory distribution following optimization using all 3 constraints: the collision, position, and orientation likelihoods, as explained earlier. 
Through sufficient weight on the grasping term, the resulting end effector alignment enables the robot to pick up the can. \Cref{fig:graspingexample_coll}, \Cref{fig:graspingexample_pos} and \Cref{fig:graspingexample_or} show the resulting distribution in the absence of the collision, position and orientation terms, respectively. In all 3 cases, grasping is unsuccessful. Without collision information, the grasping constraints compete against the KL divergence, and the position constraint is not enforced completely. Without position information, the trajectory is collision-free and the end effector is aligned, but is located away from the table. 
Finally, without orientation, the end effector is not aligned for grasping. Note that in all examples, the uncertainty contracts and then expands again. 
This is a consequence of predicting at timesteps not seen during training, that is, timesteps which are either in-between, before, or after the waypoints. 
Note that this behavior may not be present in other GP-related motion planners, particularly if they implement predictions of this form using interpolation algorithms used atop the planner's output \cite{gpmp2, gvi, sgpmp} rather than using the uncertainty obtained from the Gaussian process.

\end{document}